%% file: neurips_2026.tex
\documentclass{article}
\input{preamble}

\title{
    Camera3R: Feed-Forward 3D Reconstruction
    \\
    from Heterogeneous Camera Models
}
\author{
    Ruxiao Duan\textsuperscript{1,2} \quad
    Yunwen Zhou\textsuperscript{2} \quad
    Erin Hong\textsuperscript{2} \quad
    Dongxu Zhao\textsuperscript{2}
    \\
    \textbf{
    Eric Turner\textsuperscript{2} \quad
    Federico Tombari\textsuperscript{2} \quad
    Alex Wong\textsuperscript{1}
    }
    \vspace{3mm}
    \\
    \textsuperscript{1}Yale University \quad
    \textsuperscript{2}Google
    \\
    \tt\small \textsuperscript{1}\{ruxiao.duan, alex.wong\}@yale.edu
    \\
    \tt\small \textsuperscript{2}\{ruxiao, verse, honge, dongxuz, elturner, tombari\}@google.com
}

\begin{document}
\maketitle

\begin{abstract}
    Feed-forward 3D reconstruction (3R) from unconstrained multi-view images has recently emerged as a prominent research direction in 3D computer vision.
    Various foundational reconstruction models have been proposed with different architectures and reconstruction paradigms.
    Despite being trained on large-scale perspective datasets with substantial computational resources, these models generally struggle to generalize to non-perspective images.
    This limitation hinders their deployment in applications involving arbitrary, unknown, and potentially heterogeneous camera models.
    To build on the capabilities of existing reconstruction models with minimal retraining, we propose Camera3R, a low-rank adaptation framework that efficiently extends perspective-pretrained 3R networks to any other camera models.
    Extensive experiments demonstrate that, with only $0.5\%$ additional parameters and less than a day of training, Camera3R adapts a foundational 3D reconstruction model to non-perspective cameras while fully preserving its original performance on perspective inputs.
    Code is available at https://github.com/android-xr/camera3r.
\end{abstract}

\section{Introduction}

Three-dimensional (3D) visual mapping of an environment is a fundamental component in many spatial applications, ranging from autonomous navigation to extended reality systems. Decision-making within these applications rests on the fidelity of the map, e.g., accuracy of estimated points (and position and orientation of the platform), as well as completeness, or coverage, of the environment as observed by a set of cameras. In practice, the two metrics are not independently optimizable, but are coupled through sensing trade-offs. For a fixed sensing budget, a reduction in field-of-view (FoV) increases the angular resolution, which often results in higher fidelity estimates. On the other hand, an increase in FoV naturally improves environmental coverage, but typically introduces non-trivial lens distortion and measurement noise that degrade per-point accuracy. This trade-off motivates the use of multiple cameras, e.g., combinations of perspective and fisheye or panoramic (360$^\circ$) cameras, within a mixed-camera system whose cameras provide complementary fidelity and coverage.

While mixed-camera systems offer the potential to balance map fidelity and environmental coverage in 3D reconstruction, existing feed-forward 3D reconstruction models~\cite{wang2025vggt,wang2026pi,keetha2026mapanything} cannot fully exploit the non-perspective or heterogeneous camera inputs. Despite being trained on large-scale datasets comprising tens of millions of images, these models cannot generalize to images captured by wide-FoV or mixed cameras. This failure mode can be attributed to a covariate shift between the training and deployment imaging platforms: The training data used by current feed-forward reconstruction models is predominantly composed of perspective images. In contrast, images captured by different camera models exhibit varying fields of view, resolutions, and distortion profiles, leading to heterogeneous photometric and spatial sampling statistics. Consequently, these models do not directly transfer to mixed-camera streams, limiting their potential to provide high-fidelity maps with broad environmental coverage, despite their strong generalization across diverse 3D scenes.

A natural response to this limitation is to train separate 3D reconstruction models tailored to each camera type or imaging configuration, and subsequently register their outputs and combine them into a single map. However, such an approach would incur substantial operational overhead, and is poorly suited to spatial applications, which are often constrained by compute, memory, and overall system complexity. A key insight is that, despite differences in photometric and spatial sampling statistics due to the image formation process, images of the same 3D scene captured by different camera models should correspond to a single consistent environmental map. In other words, the heterogeneity in camera models may alter the observations, but not the underlying 3D structures being reconstructed. This intuition motivates our central thesis of whether a single feed-forward 3D reconstruction model trained primarily on images captured by perspective cameras can be systematically extended to operate on heterogeneous camera inputs without retraining from scratch.

\input{fig/teaser}

To this end, we introduce Camera3R, a framework that enables a single feed-forward model to ingest images captured by multiple camera types to produce a consistent map. The crux of our method lies in selective adaptation to inputs during inference time through a mixture of experts. We assume access to a pretrained feed-forward model and that images can originate from a broad categorization of perspective, fisheye, and panoramic cameras. We associate each camera model with an expert, each parameterized by a set of weights. We begin by freezing the pretrained model; as it is pretrained on large-scale datasets of perspective images, we consider it an expert in itself. We surmise that the pretrained model has already learned sufficiently expressive features to model diverse 3D scenes and that the same features can be efficiently modulated, rather than relearned, to extend the pretrained model to accommodate alternative camera types. Accordingly, we introduce low-rank adaptation (LoRA)~\cite{hu2022lora} to the network layers to yield experts for fisheye and panoramic cameras. However, it remains a question as to how much capacity should be allocated to each camera type to model specific distortions and viewing statistics. To address this, we employ a rank‑truncation scheme that redistributes capacity by allocating higher ranks to layer-expert pairs based on need, allowing us to leverage the existing knowledge to its full extent while adding minimal overhead in parameters to the model. The result is an efficient model unified across cameras that can be trained with an order of magnitude less data than the base model itself. 

In most spatial applications, the origin of the camera data stream is known; therefore, expert selection based on camera identity is sufficient to adaptively run inference on each image. Nonetheless, to accommodate general usage, we propose to learn a camera router which determines the expert to use for modulating latent features. Hence, at test time, we do not assume access to camera metadata and can handle inputs of any single camera type, as well as a collection of images originating from a mixture of heterogeneous cameras. This design naturally enables backward compatibility, where the generalizability and performance of existing feed-forward 3D reconstruction models are preserved. At the same time, new cameras can be introduced into an existing platform to improve both map fidelity and environmental coverage, and be accommodated without disruption to established workflows. We evaluate our method on multiple 3D reconstruction datasets, spanning indoor and outdoor environments. In single‑camera settings, our method achieves a significant improvement over the baseline methods and fully preserves performance on perspective images. We further demonstrate that, in mixed‑camera scenarios, our approach can handle a collection of images originating from any of the cameras from a heterogeneous set without prior knowledge of their identity to yield consistent and high-fidelity reconstructions. Our method is applied to three different feed-forward 3D reconstruction models to demonstrate its generalizability. 

Our contributions are as follows: 
(1) We extend existing feed-forward 3D reconstruction models pretrained on perspective imagery to operate on fisheye and panoramic camera inputs through a mixture‑of‑experts framework with adaptive routing.
(2) We introduce an adaptive capacity allocation mechanism that selectively allocates LoRA‑based adaptation capacity across experts and layers, significantly reducing the parameter overhead.
(3) We achieve state‑of‑the‑art performance in both single‑camera and mixed‑camera reconstruction settings across multiple indoor and outdoor datasets.
(4) To the best of our knowledge, we are the first to enable joint feed-forward 3D reconstruction from heterogeneous perspective, fisheye, and panoramic images in a single forward pass.

\section{Related Work}

\paragraphtitle{Feed-Forward 3D Reconstruction.}
Recent progress in 3D reconstruction has been strongly impacted by the introduction of unified, feed-forward foundational models that infer camera poses and scene geometry from unconstrained images in a single forward pass.
The trend can be traced back to DUSt3R~\cite{wang2024dust3r}, which directly predicts dense point maps from image pairs.
MASt3R~\cite{leroy2024grounding} further advances the framework by improving reconstruction quality through explicit feature matching.
Spann3R~\cite{wang20253d} subsequently extends this pairwise paradigm to image collections by aggregating information with spatial memory and directly regressing global point maps without optimization-based global alignment.
VGGT~\cite{wang2025vggt} pioneered a unified, general-purpose framework that predicts camera parameters, depth maps, point maps, and point tracks from multiple images in a single forward pass with an alternating-attention transformer architecture.
$\pi^3$~\cite{wang2026pi} introduces a permutation-equivariant formulation that eliminates performance dependence on a fixed reference view and input order.
MapAnything~\cite{keetha2026mapanything} incorporates optional geometric inputs for flexible metric reconstruction.
These works collectively mark a shift towards unified, feed-forward visual-geometry foundational models for general-purpose 3D reconstruction.

\paragraphtitle{Feed-Forward 3D Reconstruction for Non-Perspective Images.}
Despite their strong performance on standard perspective imagery, these foundational reconstruction models generally fail on non-perspective inputs.
Some recent works have developed reconstruction models for non-perspective images.
PanoVGGT~\cite{guo2026panovggt} and Argus~\cite{li2026argus} achieve feed-forward reconstruction from panoramic images based on VGGT.
Wid3R~\cite{jung2026wid3r} employs spherical harmonics for ray representations to handle perspective, fisheye, and panoramic inputs in reconstruction models, but does not demonstrate reconstructions from mixed-camera sequences.
These approaches either specialize in a particular camera model or rely on known camera types during inference.
Furthermore, they do not explicitly address mixed-camera input sequences or guarantee backward compatibility on perspective images.
CAM3R~\cite{guruprasad2026cam3r} extends the pairwise reconstruction to camera-agnostic settings by predicting per-pixel rays and radial distances, using a ray-aware global alignment stage to combine pairwise predictions across views.
Fisheye3R~\cite{duan2026fisheye3r} formulates the task of generalizing reconstruction models to fisheye cameras as an adaptation problem and introduces calibration tokens \cite{gangopadhyay2025extending} to extend perspective-pretrained reconstruction networks~\cite{wang2025vggt,wang2026pi,keetha2026mapanything} to fisheye cameras without performance degradation on perspective inputs, but the scope is limited to the fisheye domain.
In contrast, our work extends foundational multi-view reconstruction models to handle perspective, fisheye, and panoramic inputs simultaneously in a single forward propagation, accepting sequences with unknown and heterogeneous camera models, while ensuring backward compatibility on perspective data.

\paragraphtitle{Low-Rank Adaptation and Adaptive Rank Allocation.}
Low-rank adaptation (LoRA)~\cite{hu2022lora} is a parameter-efficient finetuning strategy originally proposed to adapt large language models.
Instead of updating the entire set of model parameters, LoRA freezes the backbone and injects lightweight learnable low-rank matrices into selected layers for efficient adaptation, reducing the number of trainable parameters while preserving the original model's knowledge.
Recent works further combine LoRA with mixture-of-experts routing~\cite{wu2024mixture,li2025lora,zhang2025more}, where multiple low-rank experts are selected or combined to improve specialization across inputs or tasks.
To adaptively allocate resources and achieve lightweight adaptation, adaptive-rank and post-hoc compression approaches such as~\cite{zhang2023adalora,kumaravelu2026post,vulic2026lora} have been introduced.

\section{Method}

\input{fig/method}

We study the problem of extending multi-view 3D reconstruction models pretrained on perspective imagery to handle sequences of arbitrary, unknown, and heterogeneous camera types during inference.
Three camera models are considered specifically, including perspective, fisheye, and panoramic, due to their relative prevalence in real-world 3D reconstruction applications.

\subsection{Feed-Forward Multi-View 3D Reconstruction}

Given a set of $V$ unconstrained RGB images $\{I_i\}_{i=1}^V$ describing a 3D scene, where the $i$-th frame $I_i \in [0, 1]^{H\times W \times 3}$ for each $i \in [V] = \{1, \dots, V\}$, we denote the camera model of $I_i$ as $c_i \in \mathcal{C} = \{c_\mathrm{persp}, c_\mathrm{fish}, c_\mathrm{pano}\}$.
Typical 3D reconstruction methods~\cite{wang2025vggt,wang2026pi,keetha2026mapanything} assume all the images are perspective, i.e., $c_i = c_\mathrm{persp}$ for all $i$, and process an input sequence by three steps in general.

First, an image encoder $f_\mathrm{img}$, usually a pretrained frozen DINOv2~\cite{oquab2023dinov2}, extracts features from each frame separately.
An image $I_i$ is divided into $N = HW/p^2$ patches of size $p \times p$ pixels, projected to embeddings $\{ \boldsymbol{z}_{i,j}^{(0)} \}_{j=1}^N$ where $\boldsymbol{z}_{i,j}^{(0)} \in \R^d$ is the $j$-th patch embedding of the $i$-th frame and $d$ is the embedding dimension, and then concatenated along with a class token $\boldsymbol{z}_{i,0}^{(0)} = \boldsymbol{\phi}_\mathrm{cls} \in \R^d$ as
\begin{equation}
    \boldsymbol{z}_i^{(0)} = \left[
        \boldsymbol{z}_{i,j}^{(0)}
    \right]_{j=0}^N \in \R^{(N+1) \times d},
    \quad
    \forall i \in [V].
\end{equation}
For an encoder $f_\mathrm{img}$ with $L$ transformer blocks, the $\ell$-th block processes the frame embeddings as
\begin{equation}
    \label{eq:encoder}
    \boldsymbol{z}_i^{(\ell)} = f_\mathrm{img}^{(\ell)} \left( \boldsymbol{z}_i^{(\ell-1)} \right) \in \R^{(N+1) \times d},
    \quad
    \forall i \in [V],
    \;
    \forall \ell \in [L],
\end{equation}
by multi-head self-attention and multi-layer perceptrons (MLPs)~\cite{dosovitskiy2020image}.

Second, the encoded frame features of the entire sequence are passed to a sequence encoder $f_\mathrm{seq}$, which typically employs an alternating attention architecture~\cite{wang2025vggt}, to jointly extract sequence features for 3D reconstruction as
\begin{equation}
    \left[ \boldsymbol{z}_i \right]_{i=1}^V = f_\mathrm{seq} \left(
        \left[ \boldsymbol{z}_i^{(L)} \right]_{i=1}^V
    \right) \in \R^{V \times (N+1) \times d},
\end{equation}
where $\boldsymbol{z}_i \in \R^{(N+1) \times d}$ is the final features of the $i$-th frame after the entire encoding process.

Finally, the features of each frame are forwarded to a decoder $f_\mathrm{dec}$ to predict the camera pose and dense scene attributes as
\begin{equation}
    \left( \boldsymbol{q}_i, \boldsymbol{t}_i \right), \mathcal{G}_i = f_\mathrm{dec}(\boldsymbol{z}_i),
    \quad
    \forall i \in [V],
\end{equation}
where $\boldsymbol{q}_i \in \R^4$ and $\boldsymbol{t}_i \in \R^3$ represent the predicted camera-to-world rotation quaternion and translation of frame $i$, and $\mathcal{G}_i \in \R^{H \times W \times D}$ denotes the pixel-wise dense predictions concatenated to $D$ output channels, possibly including depth maps, point maps, ray directions, or other geometric quantities, depending on the model design.

\subsection{Extension to Non-Perspective Camera Models}

\paragraphtitle{Low-rank adaptation to arbitrary cameras.}
An overview of our approach is provided in~\Cref{fig:method}.
For a reconstruction model with $K$ linear layers, suppose the $k$-th linear layer has a weight matrix $W_k \in \R^{n_k \times m_k}$ where $m_k$ and $n_k$ denote the layer's input and output dimensions.
The layer derives
\begin{equation}
    \boldsymbol{y}_{i,k} = W_k \boldsymbol{x}_{i,k},
    \quad
    \forall i \in [V],
    \;
    \forall k \in [K],
\end{equation}
where $\boldsymbol{x}_{i,k} \in \R^{m_k}$ is an input vector from frame $i$ sent to layer $k$ (with the frame's token index omitted for simplicity) and $\boldsymbol{y}_{i,k} \in \R^{n_k}$ is the output vector.
To extend the reconstruction model to non-perspective images while preserving its performance on perspective inputs, each layer is adapted by camera-specific LoRA experts as
\begin{equation}
    \label{eq:molora}
    \boldsymbol{y}_{i,k}
    = W_k \boldsymbol{x}_{i,k} + \sum_{c \in \mathcal{C} \setminus \{c_\mathrm{persp}\} } \mathbbm{1}_{c = c_i} B_k^c A_k^c \boldsymbol{x}_{i,k},
    \quad
    \forall i \in [V],
    \;
    \forall k \in [K],
\end{equation}
where $B_k^c \in \R^{n_k \times r_k^c}$ and $A_k^c \in \R^{r_k^c \times m_k}$ are learnable low-rank matrices with rank $r_k^c \leq \min(m_k, n_k)$ that are injected into the layer for adaptation to a non-perspective camera $c$, and $\mathbbm{1}$ is the indicator function to activate the corresponding camera expert for adaptation.

With this formulation, each non-perspective camera LoRA expert specializes in adapting the layer to that specific camera's projection geometry, while perspective frames bypass all these adaptations and are processed by the original weights, thereby preserving the model's prior performance.
During training, the original weights $W_k$ are frozen and the LoRA matrices are optimized, with random Gaussian initialization for each $A_k^c$ and zero for each $B_k^c$'s~\cite{hu2022lora}.
Although this paper focuses on perspective, fisheye, and panoramic images, this framework naturally extends to other camera models by introducing additional experts.

\paragraphtitle{Input-conditioned adaptation for unknown cameras.}
In~\Cref{eq:molora}, the camera type of each frame $c_i$ must be known to obtain the binary indicator $\mathbbm{1}_{c_i = c}$.
However, in real-world scenarios, camera metadata is often unavailable at test time.
To generalize the framework to handle potentially unknown cameras during inference, we modify each layer as
\begin{equation}
    \boldsymbol{y}_{i,k} = 
    \begin{cases}
        W_k \boldsymbol{x}_{i,k} + \sum_{c \in \mathcal{C} \setminus \{c_\mathrm{persp}\} } \mathbbm{1}_{c = c_i} B_k^c A_k^c \boldsymbol{x}_{i,k},
        & \!\!\!\text{if $c_i$ known},
        \\
        W_k \boldsymbol{x}_{i,k} + \sum_{c \in \mathcal{C} \setminus \{c_\mathrm{persp}\} } g_c\left( I_i \right) B_k^c A_k^c \boldsymbol{x}_{i,k},
        & \!\!\!\text{if $c_i$ unknown},
    \end{cases}
    \,
    \forall i \in [V],
    \,
    \forall k \in [K],
\end{equation}
where $I_i$ is the $i$-th frame.
The function $g: \R^{H \times W \times 3} \rightarrow \{0, 1\}^{|\mathcal{C}|}$ is a LoRA router that predicts a one-hot camera vector for each frame, and $g_c(\cdot) \in \{0, 1\}$ denotes the binary prediction of $g(\cdot)$ on camera $c$.
This discrete routing strategy that selects a single LoRA expert is employed instead of the soft mixture over experts in order to let each expert focus solely on adapting to images of that specific camera projection model, to prevent leakage of adaptation to perspective frames, and to facilitate future introduction of new camera experts.

\paragraphtitle{Incremental learning with novel cameras.}
A prospective advantage of our camera-specific LoRA experts is its natural feasibility for continual learning.
In standard full finetuning and single-expert adaptation design, incorporating a new camera sensor inevitably risks catastrophic forgetting of earlier knowledge on existing camera models.
In contrast, as our framework freezes the underlying foundational backbone and encapsulates camera-specific geometric modulations into isolated LoRA experts, new camera models (e.g., catadioptric or microscopic) can be incorporated incrementally.
Specifically, supporting a novel camera model requires only instantiating a new lightweight LoRA expert and extending the camera predictor by one classification category, without full retraining of the perspective backbone or existing adapters.

\paragraphtitle{Post-hoc model compression.}
Different camera models exhibit different distortion characteristics and may require different levels of adaptation.
Moreover, the adaptation effect of LoRA can vary substantially across layers depending on the role of each layer in the specific reconstruction model.
Therefore, instead of uniformly assigning the same rank to every layer and camera expert or using expensive hyperparameter search to obtain the sufficient rank for each layer-expert pair, we apply a post-hoc adaptive rank truncation scheme that automatically allocates ranks to more important layer-expert pairs.
Specifically, the camera adapters across all layers are initially trained with a moderately high rank, obtaining matrices $(A_k^c, B_k^c)$ for each layer $k$ and non-perspective camera $c$.
Then, singular value decomposition (SVD) is performed on the learned update matrix as
\begin{equation}
    B_k^c A_k^c = U_{k,c} \Sigma_{k,c} V_{k,c}^\top,
    \quad
    \forall k \in [K],
    \;
    \forall c \in \mathcal{C} \setminus \{c_\mathrm{persp}\},
\end{equation}
where $U_{k,c} \in \R^{n_k \times r_k^c}$, $V_{k,c} \in \R^{m_k \times r_k^c}$, and $\Sigma_{k,c} = \mathrm{Diag}(\boldsymbol{\sigma}_{k,c}) \in \R^{r_k^c \times r_k^c}$ is a diagonal matrix with a diagonal $\boldsymbol{\sigma}_{k,c} \in \R^{r_k^c}$ that contains the singular values in descending order.
According to the Eckart-Young-Mirsky theorem~\cite{eckart1936approximation}, truncated SVD yields the optimal low-rank matrix approximation under unitarily invariant norms.
Leveraging this theoretical guarantee, we apply a global threshold $\sigma_0$ across all layers and camera models to truncate singular values below it as
\begin{equation}
    U'_{k,c} \Sigma'_{k,c} V_{k,c}^{'\top}
    = \left(U'_{k,c} \Sigma_{k,c}'^{1/2}\right) \left(\Sigma_{k,c}'^{1/2} V_{k,c}^{'\top}\right)
    = B_k'^{c} A_k'^{c}
\end{equation}
where $U'_{k,c} \in \R^{n_k \times r_k'^c}$, $V'_{k,c} \in \R^{m_k \times r_k'^c}$, and $\Sigma'_{k,c} = \mathrm{Diag}(\boldsymbol{\sigma}'_{k,c}) \in \R^{r_k'^c \times r_k'^c}$ are the truncated SVD matrices, $\boldsymbol{\sigma}'_{k,c} \in \R^{r_k'^c}$ contains the first $r_k'^c$ singular values of $\boldsymbol{\sigma}_{k,c}$ that reach the threshold $\sigma_0$, $r_k'^c$ is the truncated rank, and $B_k'^{c} \in \R^{n_k \times r_k'^c}$ and $A_k'^{c} \in \R^{r_k'^c \times m_k}$ are the truncated LoRA matrices.
The global threshold $\sigma_0$ is chosen such that LoRA experts consume a predetermined parameter budget.

\input{fig/augmentation}

\subsection{Learning}

\paragraphtitle{Data augmentation.}
As demonstrated in~\Cref{fig:augmentation}, we adapt the models on multi-view 3D fisheye and panoramic datasets, and further augment them through cross-camera synthesis.
It is unnecessary to introduce perspective datasets for training which can be synthesized from wide-FoV data.
Specifically, from fisheye sequences, we randomly distort the images with varying distortion parameters to synthesize diverse distortions, and we randomly undistort the images to obtain perspective views to enable cross-camera learning \cite{gangopadhyay2025extending,gangopadhyay2026from}.
From panoramic sequences, we apply random rotation first and then render with both perspective and fisheye projections from the panoramic view.
These transformations significantly enrich the diversity of camera distortions in the training set and provide multiple camera projections of the same underlying scene, facilitating effective learning of camera-dependent features and improving the robustness of the camera predictor.

\paragraphtitle{Model training.}
The camera classifier is trained beforehand with cross-entropy loss based on the ground-truth camera labels.
The individual LoRA experts are firstly trained with fully fisheye and panoramic sequences separately.
Global attention LoRA parameters are subsequently finetuned on the mixed-camera sequences.
Adaptation is first performed with a rank of $64$ for all layers, after which post-hoc compression is applied and the LoRA matrices are compressed to consume only $\epsilon = 0.5\%$ of backbone parameters.
For LoRA modules, we use their ground-truth camera labels for routing to train the weight matrices, and the training objective is simply the 3D reconstruction loss used by the chosen backbone model.
During inference, the router is replaced by the camera predictor we trained to deal with unknown cameras.

\section{Experiments}

\subsection{Settings}

\paragraphtitle{Datasets.}
We train models on three fisheye datasets (ASE~\cite{engel2023project}, ScanNet++~\cite{yeshwanth2023scannet++}, and KITTI360~\cite{liao2022kitti}) and four panoramic datasets (Matterport3D~\cite{chang2017matterport3d}, Gibson~\cite{xia2018gibson}, TartanAirV2~\cite{wang2020tartanair}, and PanoCity~\cite{guo2026panovggt}).
The models are evaluated on four camera settings.
Perspective: undistorted ADT~\cite{pan2023aria} and ETH3D~\cite{schops2017multi}.
Fisheye: ADT~\cite{pan2023aria} and ScanNet++~\cite{yeshwanth2023scannet++}.
Panoramic: Stanford2D3DS~\cite{armeni2017joint} and 360Loc~\cite{huang2024360loc}.
Heterogeneous: KITTI360~\cite{liao2022kitti} with mixed perspective and fisheye images, and Stanford2D3DS~\cite{armeni2017joint} and 360Loc~\cite{huang2024360loc} with mixed perspective, fisheye, and panoramic images.

\paragraphtitle{Models.}
We test the effectiveness of adaptation on three foundational backbone models: VGGT~\cite{wang2025vggt}, MapAnything~\cite{keetha2026mapanything}, and $\pi^3$~\cite{wang2026pi}.
We use the latest MapAnything v1.1 and $\pi^3$-X checkpoints.

\paragraphtitle{Baselines.}
We compare our performance with previous foundational reconstruction models for non-perspective images, including Wid3R~\cite{jung2026wid3r}, Fisheye3R~\cite{duan2026fisheye3r}, PanoVGGT~\cite{guo2026panovggt} (with a low-resolution version and a high-resolution one denoted as PanoVGGT-H), and Argus~\cite{li2026argus}.
We additionally compare with calibration tokens (CT) introduced in~\cite{duan2026fisheye3r} as a baseline of adaptation algorithm, applied in the same models and trained with the same data and loss for an apples-to-apples comparison.

\paragraphtitle{Metrics.}
We evaluate the models from four different tasks: camera pose, depth map, point map, and FoV map estimation, following~\cite{wang2024dust3r,wang2025vggt,wang2026pi,keetha2026mapanything,duan2026fisheye3r}.
For camera pose, we adopt the Relative Rotation Accuracy (RRA), Relative Translation Accuracy (RTA), and Area Under Curve (AUC) of the accuracy-threshold curve at 5$^\circ$ for angular accuracy and Absolute Trajectory Error (ATE), Relative Pose Error for translation (RPEt), and Relative Pose Error for rotation (RPEr) for distance error quantification.
For depth map estimation, Absolute Relative Error (Rel), Root Mean Squared Error (RMSE), and $\delta_1$ accuracy are selected as metrics.
For point map estimation, Accuracy (Acc), Completeness (Comp), Chamfer Distance (CD), and Normal Consistency (NC) are chosen.
For FoV map evaluation, the median error for horizontal ($\text{hErr}$) and vertical ($\text{vErr}$) FoVs in degrees, along with the Recall (Rec) at a $1^\circ$ threshold for FoV are measured.
The arrow $\uparrow$ behind a metric means the higher the better, while $\downarrow$ means the lower the better.

\input{tab/quantitative_fisheye}
\input{tab/quantitative_panoramic}

\subsection{Results}

\paragraphtitle{Camera router training and classification accuracy.}
The camera model predictor is implemented as a lightweight convolutional neural network.
With our training setup, the classifier converges in a few minutes and achieves $100\%$ classification accuracy on our test sets.
This observation suggests that the distortion characteristics of the three camera models are sufficiently distinct from their images.

\paragraphtitle{Quantitative results.}
We compare with different methods on fisheye and panoramic sequences in~\Cref{tab:quantitative_fisheye,tab:quantitative_panoramic} and mixed-camera sequences in~\Cref{tab:quantitative_mixed}.
Results across all these datasets and evaluation metrics demonstrate that our approach significantly beats prior ones by a large margin, with the strongest performance on $\pi^3$ backbone.
As an adaptation method, our method can be applied to all the three backbones, extending all of them to non-perspective images, with a much better effect than calibration tokens.
\Cref{tab:quantitative_experts} compares several design choices for handling inputs with heterogeneous camera models.
The notation for different LoRA expert assignments is defined in the table caption.
Compared with the original model, both full finetuning and retraining degrade performance on perspective inputs, as the original backbone weights are no longer preserved.
Among the LoRA variants, assigning an adaptation expert to perspective images, as in 3PFP and 1PFP, also leads to worse perspective performance, likely because learning a new perspective expert from a relatively small dataset causes forgetting of the knowledge acquired by the backbone during large-scale pretraining.
In contrast, removing LoRA expert for perspective images, as in 1FP and 2FP, perfectly preserves the original perspective performance by routing perspective inputs through the frozen backbone.
Compared to 1FP, 2FP generally achieves superior performance in fully fisheye and panoramic images, indicating that separating adaptation of different cameras to different experts is a better design.
After rank truncation, the adapter is approximately compressed by 22 times and the performance drops only slightly, presenting an efficiency-performance trade-off.

\input{tab/quantitative_mixed}
\input{tab/quantitative_experts}

\paragraphtitle{Qualitative results.}
Qualitative figures of how our method adapts models to fisheye and panoramic sequences are presented in~\Cref{fig:vis_scannetpp,fig:vis_adt,fig:vis_stanford2d3ds,fig:vis_360loc}.
\Cref{fig:qualitative} illustrates two scenarios of applying our approach to $\pi^3$ for mixed cameras, with the predicted point clouds before and after adaptation.
In the KITTI360 example, two fisheye cameras are mounted back-to-back and connected through a front-facing perspective camera.
Without support for mixed-camera inputs, a fisheye specialist model would struggle in this setting due to the lack of covisibility between the two fisheye views.
In contrast, the perspective view provides an intermediate observation that links the two fisheye views, enabling a coherent reconstruction of the entire scene.
In the 360Loc example, our model is also able to accurately reconstruct the scene given three completely distinct camera models in the inputs.

\paragraphtitle{Limitations.}
Our method assumes that each input image belongs to a predefined set of camera categories, currently including perspective, fisheye, and panoramic.
Thus when an image from a fourth camera model is encountered, the classifier has to incrementally learn the new camera type before correctly routing the images to a new expert.
Additionally, our training employs cross-camera synthesis from fisheye and panoramic datasets.
Although these augmentations provide diverse projection geometries, synthetic distortion may not fully capture real lens- and sensor-specific effects.

\input{fig/qualitative}

\section{Conclusion}

We present a parameter-efficient low-rank adaptation framework to extend 3D feed-forward reconstruction models pretrained on large-scale perspective datasets to non-perspective, potentially unknown, and mixed-camera sequences.
By freezing the original reconstruction backbone and introducing lightweight learnable camera-specific LoRA experts for fisheye and panoramic images that are selected by a camera router, we separate the adaptation effect on different camera models, enabling the network to handle hybrid inputs while preserving backward compatibility with perspective data.
This design allows existing large-scale reconstruction models to be reused with minimal additional parameters and training cost, rather than requiring model-specific retraining or separate networks for each camera configuration.
We demonstrate that the framework supports mixed-camera sequences, where images from different projection models jointly contribute to a coherent reconstruction.
Across three foundational backbones, our method achieves state-of-the-art performance on 3D scene reconstruction with non-perspective images, while maintaining the original model behavior on perspective inputs.
We hope this work provides a practical step toward deploying feed-forward 3D reconstruction models in real-world systems with heterogeneous and evolving camera setups.

\newpage

\section*{Acknowledgments}
This work is supported by Google Gift Funding and NSF-2112562 Athena AI Institute.
Alex Wong was supported by the grant SFI-AN-NC-SCN-00007276 from the Simons Foundation International as part of the SCENE collaboration. The authors have no competing interests to disclose.

{
    \small
    \bibliographystyle{plainnat}
    \bibliography{references}
}

\newpage
\appendix

\section{Training and Evaluation Details}
\label{sec:implementation}

\input{tab/datasets}
\paragraphtitle{Datasets and train-test split.}
We list the dataset information in~\Cref{tab:datasets}.
We train and evaluate the models on perspective, fisheye, and panoramic cameras, spanning real and synthetic, indoor and outdoor environments.
For the two datasets that appear in both training and testing, ScanNet++ follows the split in~\cite{keetha2026mapanything}, while KITTI360 follows the split in~\cite{duan2026fisheye3r}, both of which are enforced at a scene level, thus no sequence from the test set is revealed to the model during training.

\paragraphtitle{Hyperparameters.}
For adaptation, we initially train with a uniform rank of $64$ and then truncate LoRA to $\epsilon=0.5\%$ of the original backbone's parameters.

\paragraphtitle{Hardware.}
Training and testing are performed on 8 NVIDIA RTX 6000 Blackwell (48 GB) GPUs.

\paragraphtitle{Training time.}
The entire training process takes approximately 20 hours for each model.

\paragraphtitle{Data processing.}
During training, each GPU is given a maximum of 16 images, with each image resized to have a maximum edge of 518 pixels.
With a sequence length randomly sampled as $V \in [2, 16]$, the number of sequences per GPU is $\lfloor 16/V \rfloor$, leading to an effective batch size of $8 \lfloor 16/V \rfloor$ sequences per iteration.
Evaluation is conducted on a fixed sequence length of $V = 16$.

\paragraphtitle{Photometric augmentations.}
Random color jittering, gray-scale conversion, and Gaussian blur are independently applied to each image.

\paragraphtitle{Distortion sampling.}
For fisheye distortion synthesis, we follow~\cite{duan2026fisheye3r} and adopt the Kannala-Brandt camera model~\cite{kannala2006generic} with randomly sampled distortion coefficients $(k_1, k_2, k_3, k_4)$, where $k_1, k_2, k_3 \sim U(-0.5, 0.5)$ and $k_4 \sim U(-0.05, 0.05)$.
This augmentation generates diverse fisheye distortion patterns from existing training images.
For images that are originally fisheye, we retain the original distortion with probability $0.5$ and replace it with a newly sampled distortion with probability $0.5$.
For panoramic image augmentation, we always apply a random yaw rotation sampled from $(-180^\circ, 180^\circ)$.
Additionally, with a probability of $0.5$, we independently apply random pitch and roll rotations, each sampled uniformly from $(-15^\circ, 15^\circ)$.

\paragraphtitle{Camera model sampling.}
For camera model sampling during training, if a sequence is originally fisheye, it has a probability of $0.5$ to remain entirely fisheye and a probability of $0.5$ to be converted into mixed camera types, where each frame is independently sampled as either perspective or fisheye with equal probability.
If a sequence is originally panoramic, it has a probability of $0.5$ to remain entirely panoramic and a probability of $0.5$ to be converted into mixed camera types, where each frame is independently sampled from perspective, fisheye, and panoramic cameras with uniform probabilities.
During evaluation, KITTI360 uses the dataset's original perspective and fisheye images with a sampling ratio of $1\!:\!1$.
For panoramic datasets, we additionally generate perspective and fisheye views for testing, and sample the three camera types uniformly with a ratio of $1\!:\!1\!:\!1$.

\paragraphtitle{Frame sampling scheme.}
A pairwise covisibility matrix is precomputed for each scene and only views with at least $25\%$ pixels visible from another sampled view are sampled for both training and evaluation, following~\cite{keetha2026mapanything}.

\input{fig/vis_adt}
\input{fig/vis_scannetpp}
\input{fig/vis_stanford2d3ds}
\input{fig/vis_360loc}

\end{document}

%% file: preamble.tex
\PassOptionsToPackage{numbers,sort&compress}{natbib}

 \usepackage[main, final]{neurips_2026}

\usepackage[utf8]{inputenc} 
\usepackage[T1]{fontenc}    
\usepackage{hyperref}       
\usepackage{url}            
\usepackage{booktabs}       
\usepackage{amsfonts}       
\usepackage{nicefrac}       
\usepackage{microtype}      
\usepackage{xcolor}         

\newcommand{\paragraphtitle}[1]{\noindent \textbf{#1}}

\usepackage{multirow}
\usepackage{rotating}
\usepackage{array}
\usepackage{amssymb}
\usepackage{amsmath}
\usepackage{makecell}
\usepackage{mathtools}
\usepackage{cleveref}
\usepackage{bbm}
\newcommand{\R}{\mathbb{R}}

\usepackage[table]{xcolor}
\usepackage{mathtools}
\definecolor{tabred}{rgb}{1, 0.7, 0.7}
\definecolor{taborange}{rgb}{1, 0.85, 0.7}
\definecolor{tabyellow}{rgb}{1, 1, 0.7}
\definecolor{tabgray}{rgb}{0.75, 0.75, 0.75}

%% file: fig/teaser.tex
\begin{figure*}[t]
    \centering
    \includegraphics[width=\linewidth]{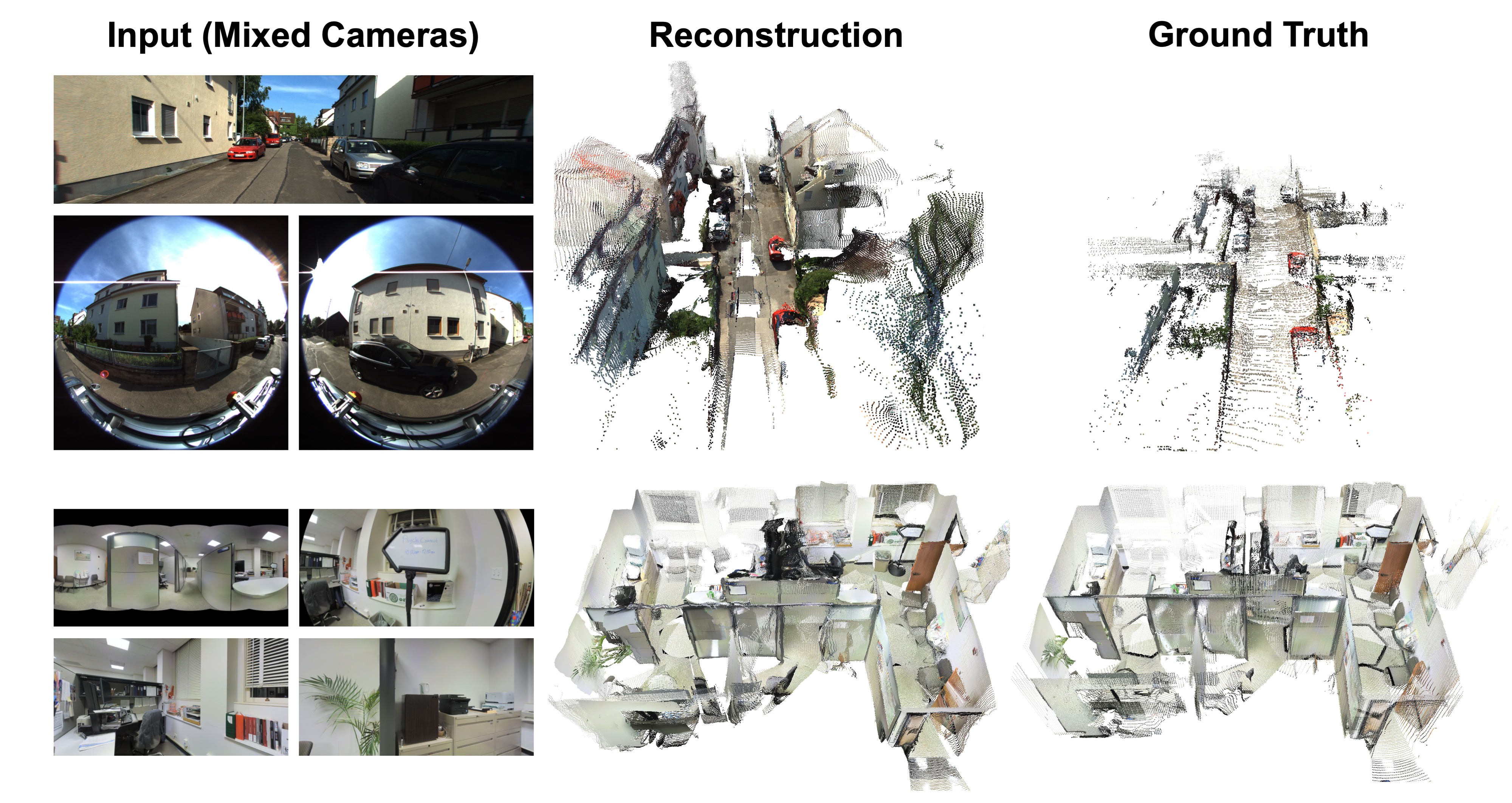}
    \vspace{-6mm}
    \caption{
        \paragraphtitle{3D reconstruction from heterogeneous cameras.}
        Given an image sequence of arbitrary, unknown, and mixed camera models, our method reconstructs the scene in a single forward pass.
    }
    \label{fig:teaser}
\end{figure*}

%% file: fig/method.tex
\begin{figure*}[t]
    \centering
    \includegraphics[width=\linewidth]{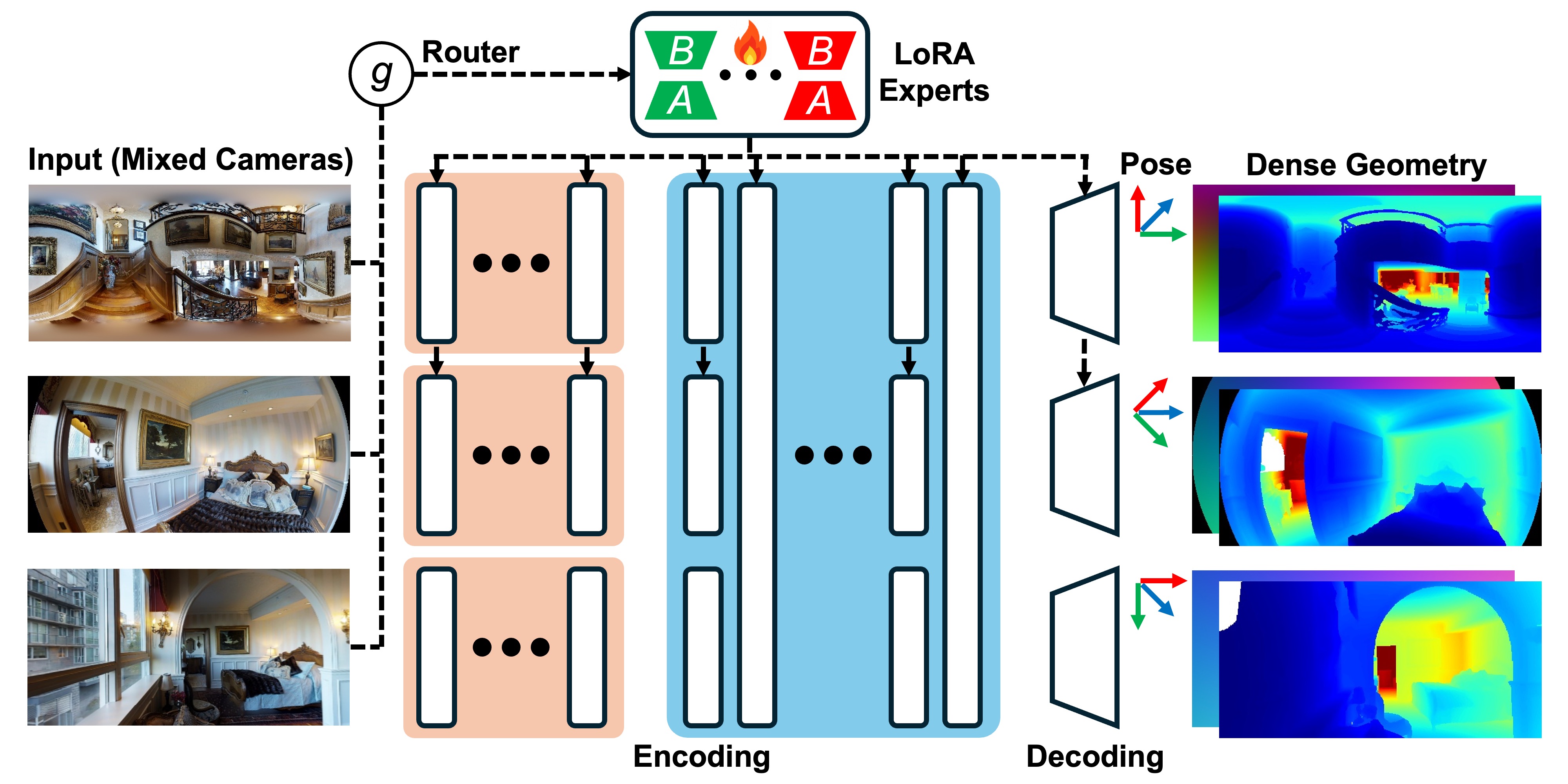}
    \vspace{-6mm}
    \caption{
        \paragraphtitle{Camera-agnostic feed-forward reconstruction.}
        Given an input sequence from arbitrary, unknown, and mixed camera models, we adapt a perspective-pretrained reconstruction network by camera-specific LoRA experts.
        Each frame is passed to a camera router, which selects the corresponding LoRA expert to adapt the model to the frame's camera geometry, enabling pose and dense geometry (depth, ray directions, etc.) prediction from non-perspective cameras.
        Perspective frames bypass the adaptation, preserving the original behavior of the pretrained backbone.
    }
    \label{fig:method}
\end{figure*}

%% file: fig/augmentation.tex
\begin{figure*}[t]
    \centering
    \includegraphics[width=\linewidth]{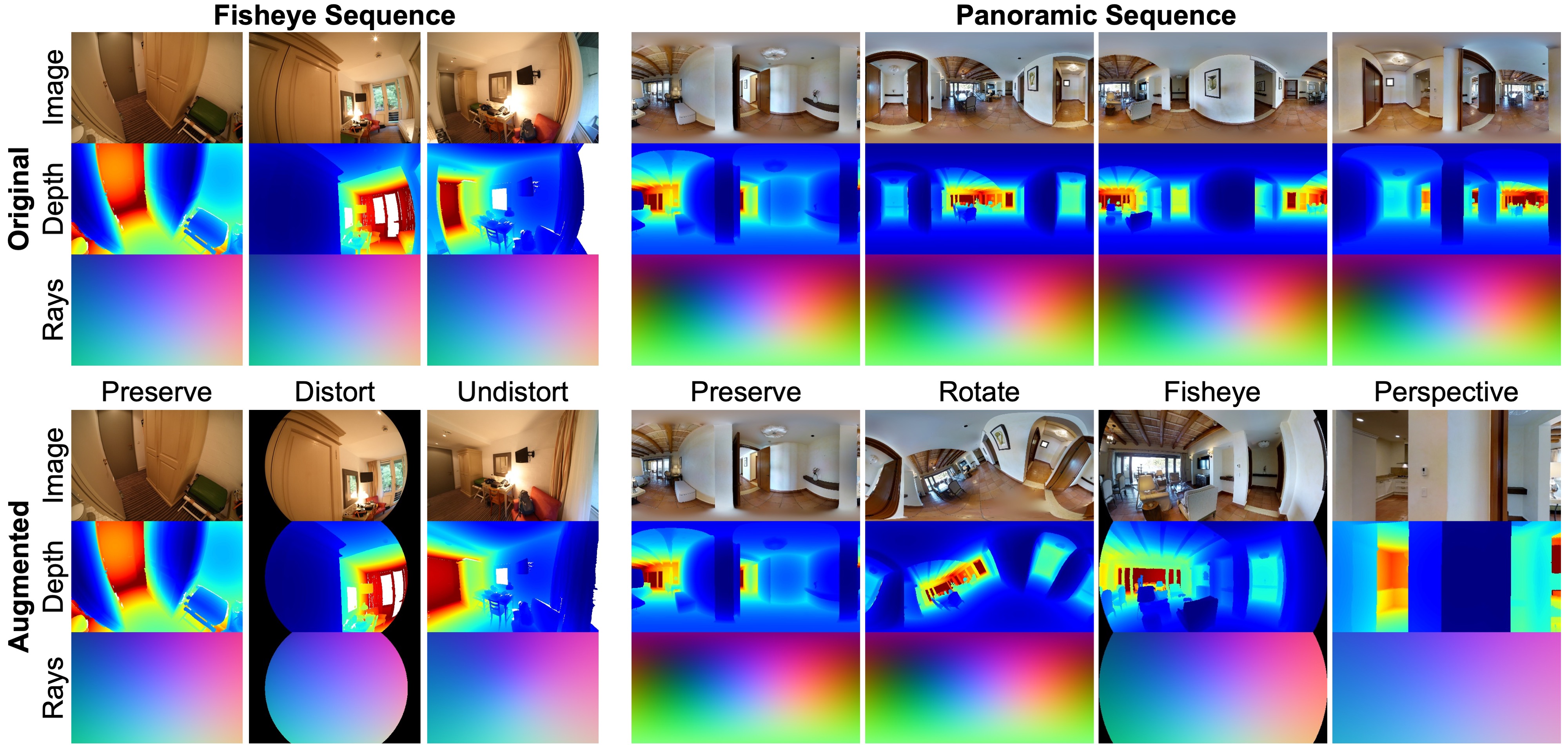}
    \vspace{-6mm}
    \caption{
        \paragraphtitle{Cross-camera data augmentation for model adaptation.}
        Multi-view 3D reconstruction data from fisheye and panoramic sequences are augmented by synthesizing diverse camera projections.
        For fisheye sequences, we randomly preserve the original views, create additional fisheye distortions, and undistort the image to obtain perspective views.
        For panoramic sequences, we randomly preserve the original views, apply rotations, and render fisheye or perspective images by sampling from the panoramic sphere.
        Dense ground truth quantities such as depth and ray maps are transformed together with the image, providing geometrically aligned supervision across various camera models.
    }
    \label{fig:augmentation}
\end{figure*}

%% file: tab/quantitative_fisheye.tex
\begin{table*}[t]
    \centering
    \resizebox{\textwidth}{!}{
        \newcolumntype{M}{>{\centering\arraybackslash}m{8mm}}
        \begin{tabular}{cl | *{4}{MMM|} MMM }
            \toprule[1pt]
            \multicolumn{2}{c|}{\multirow{2}{*}{Dataset/Method}} &
            \multicolumn{3}{c|}{Pose (Angular)} &
            \multicolumn{3}{c|}{Pose (Distance)} &
            \multicolumn{3}{c|}{Depth} &
            \multicolumn{3}{c|}{Point} &
            \multicolumn{3}{c}{FoV}
            \\
            &
            & RRA$\uparrow$ & RTA$\uparrow$ & AUC$\uparrow$
            & ATE$\downarrow$ & RPEt$\downarrow$ & RPEr$\downarrow$
            & Rel$\downarrow$ & RMSE$\downarrow$ & $\delta_1$$\uparrow$
            & Acc$\downarrow$ & Comp$\downarrow$ & NC$\uparrow$
            & hErr$\downarrow$ & vErr$\downarrow$ & Rec$\uparrow$
            \\ \midrule
            \multirow{11}{*}[-6pt]{\rotatebox{90}{ADT}}
            & Wid3R & 0.411 & 0.403 & 0.077 & 0.111 & 0.192 & 7.153 & 0.088 & 0.468 & 0.943 & 0.136 & 0.115 & 0.703 & 5.852 & 2.267 & 0.120 \\
            & Fisheye3R & 0.941 & 0.486 & 0.218 & 0.111 & 0.166 & 3.195 & 0.083 & 0.355 & 0.950 & 0.052 & 0.049 & 0.747 & 1.889 & 1.907 & 0.115 \\
            \cmidrule{2-17}
            & VGGT & 0.392 & 0.273 & 0.058 & 0.191 & 0.274 & 7.515 & 0.199 & 0.787 & 0.741 & 0.233 & 0.148 & 0.628 & 5.992 & 6.068 & 0.012 \\
            & +CT & 0.984 & 0.720 & 0.452 & 0.061 & 0.106 & 2.592 & 0.071 & 0.337 & 0.952 & 0.046 & 0.041 & 0.741 & 1.441 & 0.655 & 0.458 \\
            & +Ours & 0.985 & 0.802 & 0.502 & 0.047 & 0.093 & \textbf{2.401} & 0.069 & 0.331 & 0.954 & 0.039 & 0.040 & 0.752 & 1.330 & 0.913 & 0.472 \\
            \cmidrule{2-17}
            & MapAnything & 0.466 & 0.275 & 0.057 & 0.192 & 0.287 & 6.928 & 0.122 & 0.436 & 0.909 & 0.087 & 0.087 & 0.718 & 5.263 & 5.256 & 0.019 \\
            & +CT & 0.982 & 0.620 & 0.354 & 0.080 & 0.143 & 2.960 & 0.085 & 0.354 & 0.949 & 0.046 & 0.042 & 0.746 & 1.308 & 0.942 & 0.451 \\
            & +Ours & 0.986 & 0.679 & 0.407 & 0.065 & 0.121 & 2.634 & 0.083 & 0.347 & 0.950 & 0.041 & 0.041 & 0.750 & \textbf{1.274} & 0.762 & 0.519 \\
            \cmidrule{2-17}
            & $\pi^3$ & 0.540 & 0.248 & 0.061 & 0.200 & 0.275 & 6.216 & 0.106 & 0.425 & 0.925 & 0.099 & 0.086 & 0.739 & 5.248 & 5.205 & 0.015 \\
            & +CT & 0.971 & 0.732 & 0.437 & 0.054 & 0.103 & 2.957 & 0.082 & 0.345 & 0.953 & 0.042 & 0.045 & 0.747 & 1.498 & 1.359 & 0.392 \\
            & +Ours & \textbf{0.990} & \textbf{0.860} & \textbf{0.610} & \textbf{0.034} & \textbf{0.078} & 2.473 & \textbf{0.061} & \textbf{0.317} & \textbf{0.956} & \textbf{0.032} & \textbf{0.039} & \textbf{0.763} & 1.276 & \textbf{0.531} & \textbf{0.600} \\
            \cmidrule{1-17}
            \multirow{11}{*}[-6pt]{\rotatebox{90}{ScanNet++}}
            & Wid3R & 0.995 & 0.964 & 0.846 & 0.036 & 0.062 & 1.396 & 0.035 & 0.244 & 0.977 & 0.025 & 0.016 & 0.850 & 0.587 & 0.689 & 0.870 \\
            & Fisheye3R & 0.997 & 0.879 & 0.627 & 0.054 & 0.087 & 1.987 & 0.047 & 0.200 & 0.975 & 0.034 & 0.016 & 0.803 & 0.917 & 0.886 & 0.517 \\
            \cmidrule{2-17}
            & VGGT & 0.190 & 0.154 & 0.013 & 0.334 & 0.572 & 11.971 & 0.128 & 0.371 & 0.857 & 0.129 & 0.072 & 0.734 & 11.714 & 9.419 & 0.011 \\
            & +CT & 0.983 & 0.775 & 0.469 & 0.084 & 0.131 & 2.636 & 0.053 & 0.218 & 0.973 & 0.046 & 0.018 & 0.784 & 0.849 & 0.823 & 0.523 \\
            & +Ours & 0.996 & 0.955 & 0.830 & 0.037 & 0.058 & 1.386 & 0.036 & \textbf{0.191} & 0.978 & 0.022 & 0.013 & 0.839 & 0.349 & 0.381 & 0.927 \\
            \cmidrule{2-17}
            & MapAnything & 0.379 & 0.305 & 0.058 & 0.237 & 0.397 & 8.604 & 0.092 & 0.273 & 0.946 & 0.087 & 0.046 & 0.762 & 5.777 & 4.629 & 0.104 \\
            & +CT & 0.995 & 0.829 & 0.559 & 0.065 & 0.102 & 2.265 & 0.052 & 0.212 & 0.973 & 0.040 & 0.018 & 0.799 & 0.867 & 0.873 & 0.516 \\
            & +Ours & \textbf{0.998} & 0.901 & 0.702 & 0.048 & 0.072 & 1.668 & 0.043 & 0.204 & 0.976 & 0.026 & 0.015 & 0.829 & 0.512 & 0.501 & 0.841 \\
            \cmidrule{2-17}
            & $\pi^3$ & 0.269 & 0.198 & 0.024 & 0.263 & 0.455 & 9.330 & 0.083 & 0.260 & 0.957 & 0.091 & 0.042 & 0.779 & 8.611 & 6.720 & 0.050 \\
            & +CT & 0.995 & 0.889 & 0.664 & 0.054 & 0.084 & 1.865 & 0.052 & 0.217 & 0.975 & 0.043 & 0.018 & 0.791 & 0.913 & 0.824 & 0.533 \\
            & +Ours & 0.997 & \textbf{0.968} & \textbf{0.870} & \textbf{0.030} & \textbf{0.048} & \textbf{1.282} & \textbf{0.034} & \textbf{0.191} & \textbf{0.979} & \textbf{0.019} & \textbf{0.012} & \textbf{0.851} & \textbf{0.254} & \textbf{0.324} & \textbf{0.974} \\
            \bottomrule[1pt]
        \end{tabular}
    }
    \caption{
        Quantitative evaluation on fisheye sequences.
    }
    \label{tab:quantitative_fisheye}
\end{table*}

%% file: tab/quantitative_panoramic.tex
\begin{table*}[t]
    \centering
    \resizebox{\textwidth}{!}{
        \newcolumntype{M}{>{\centering\arraybackslash}m{8mm}}
        \begin{tabular}{cl | *{3}{MMM|} MMM }
            \toprule[1pt]
            \multicolumn{2}{c|}{\multirow{2}{*}{Dataset/Method}} &
            \multicolumn{3}{c|}{Pose (Angular)} &
            \multicolumn{3}{c|}{Pose (Distance)} &
            \multicolumn{3}{c|}{Depth} &
            \multicolumn{3}{c}{Point}
            \\
            &
            & RRA$\uparrow$ & RTA$\uparrow$ & AUC$\uparrow$
            & ATE$\downarrow$ & RPEt$\downarrow$ & RPEr$\downarrow$
            & Rel$\downarrow$ & RMSE$\downarrow$ & $\delta_1$$\uparrow$
            & Acc$\downarrow$ & Comp$\downarrow$ & NC$\uparrow$
            \\ \midrule
            \multirow{11}{*}[-8pt]{\rotatebox{90}{Stanford2D3DS}}
            & Wid3R & 0.751 & 0.720 & 0.517 & 1.258 & 1.654 & 21.361 & 0.063 & 0.621 & 0.969 & 0.216 & 0.227 & 0.747 \\
            & Argus & 0.839 & 0.795 & 0.635 & 0.918 & 0.856 & 12.049 & 0.067 & 0.505 & 0.966 & 0.196 & 0.299 & 0.773 \\
            \cmidrule{2-14}
            & VGGT & 0.082 & 0.052 & 0.004 & 3.844 & 4.623 & 81.735 & 0.470 & 1.612 & 0.381 & 0.535 & 2.182 & 0.529 \\
            & +CT & 0.457 & 0.396 & 0.120 & 1.497 & 1.874 & 19.382 & 0.090 & 0.494 & 0.954 & 0.278 & 0.265 & 0.693 \\
            & +Ours & 0.958 & 0.909 & 0.781 & 0.206 & 0.200 & 1.985 & 0.050 & 0.382 & 0.981 & 0.072 & 0.064 & 0.862 \\
            \cmidrule{2-14}
            & MapAnything & 0.085 & 0.040 & 0.002 & 4.398 & 4.242 & 81.574 & 0.681 & 4.916 & 0.250 & 0.466 & 3.418 & 0.507 \\
            & +CT & 0.764 & 0.702 & 0.340 & 0.682 & 0.743 & 6.915 & 0.101 & 0.535 & 0.938 & 0.170 & 0.115 & 0.717 \\
            & +Ours & 0.976 & 0.921 & 0.730 & 0.148 & 0.205 & 2.429 & 0.055 & 0.425 & 0.978 & 0.069 & \textbf{0.045} & 0.842 \\
            \cmidrule{2-14}
            & $\pi^3$ & 0.076 & 0.037 & 0.001 & 4.365 & 4.166 & 87.080 & 0.457 & 1.621 & 0.385 & 0.559 & 2.460 & 0.516 \\
            & +CT & 0.793 & 0.764 & 0.432 & 0.492 & 0.600 & 6.925 & 0.067 & 0.424 & 0.971 & 0.136 & 0.097 & 0.780 \\
            & +Ours & \textbf{0.978} & \textbf{0.945} & \textbf{0.857} & \textbf{0.105} & \textbf{0.144} & \textbf{1.364} & \textbf{0.043} & \textbf{0.377} & \textbf{0.982} & \textbf{0.048} & 0.053 & \textbf{0.888} \\
            \cmidrule{1-14}
            \multirow{12}{*}[-8pt]{\rotatebox{90}{360Loc}}
            & PanoVGGT & 0.453 & 0.385 & 0.090 & 1.117 & 1.279 & 7.649 & 0.297 & 3.634 & 0.700 & 0.651 & 0.349 & 0.689 \\
            & PanoVGGT-H & 0.997 & 0.996 & 0.933 & 0.087 & 0.118 & 0.459 & 0.259 & 3.484 & 0.739 & 0.418 & 0.332 & 0.765 \\
            & Argus & 0.900 & 0.866 & 0.548 & 0.890 & 1.140 & 3.642 & 0.290 & 4.982 & 0.711 & 0.842 & 0.763 & 0.677 \\
            \cmidrule{2-14}
            & VGGT & 0.261 & 0.089 & 0.010 & 4.735 & 6.193 & 38.487 & 0.532 & 5.313 & 0.290 & 2.165 & 5.092 & 0.560 \\
            & +CT & 0.878 & 0.694 & 0.321 & 0.574 & 0.852 & 2.517 & 0.327 & 4.021 & 0.720 & 0.682 & 0.332 & 0.684 \\
            & +Ours & \textbf{1.000} & 0.998 & 0.939 & 0.086 & 0.123 & 0.413 & \textbf{0.234} & \textbf{3.396} & \textbf{0.811} & 0.348 & 0.258 & 0.768 \\
            \cmidrule{2-14}
            & MapAnything & 0.224 & 0.048 & 0.006 & 5.632 & 7.345 & 48.683 & 0.688 & 13.669 & 0.239 & 2.202 & 6.454 & 0.550 \\
            & +CT & 0.945 & 0.902 & 0.531 & 0.302 & 0.446 & 1.818 & 0.377 & 4.049 & 0.691 & 0.654 & 0.312 & 0.644 \\
            & +Ours & 0.998 & 0.992 & 0.860 & 0.126 & 0.170 & 0.759 & 0.297 & 3.931 & 0.739 & 0.503 & 0.309 & 0.734 \\
            \cmidrule{2-14}
            & $\pi^3$ & 0.283 & 0.053 & 0.006 & 5.122 & 6.734 & 38.235 & 0.480 & 4.734 & 0.370 & 2.102 & 5.327 & 0.551 \\
            & +CT & 0.980 & 0.982 & 0.694 & 0.185 & 0.286 & 1.371 & 0.333 & 3.634 & 0.742 & 0.472 & \textbf{0.240} & 0.711 \\
            & +Ours & \textbf{1.000} & \textbf{0.999} & \textbf{0.969} & \textbf{0.068} & \textbf{0.099} & \textbf{0.354} & 0.272 & 3.752 & 0.790 & \textbf{0.338} & 0.265 & \textbf{0.781} \\
            \bottomrule[1pt]
        \end{tabular}
    }
    \caption{
        Quantitative evaluation on panoramic sequences.
    }
    \label{tab:quantitative_panoramic}
\end{table*}

%% file: tab/quantitative_mixed.tex
\begin{table*}[t]
    \centering
    \resizebox{\textwidth}{!}{
        \newcolumntype{M}{>{\centering\arraybackslash}m{10mm}}
        \begin{tabular}{cl | *{3}{MMM|} MMM }
            \toprule[1pt]
            \multicolumn{2}{c|}{\multirow{2}{*}{Dataset/Method}} &
            \multicolumn{3}{c|}{Pose (Angular)} &
            \multicolumn{3}{c|}{Pose (Distance)} &
            \multicolumn{3}{c|}{Depth} &
            \multicolumn{3}{c}{Point}
            \\
            &
            & RRA$\uparrow$ & RTA$\uparrow$ & AUC$\uparrow$
            & ATE$\downarrow$ & RPEt$\downarrow$ & RPEr$\downarrow$
            & Rel$\downarrow$ & RMSE$\downarrow$ & $\delta_1$$\uparrow$
            & Acc$\downarrow$ & Comp$\downarrow$ & NC$\uparrow$
            \\ \midrule
            \multirow{3}{*}{\rotatebox{90}{\scriptsize KITTI360}}
            & Wid3R & 0.255 & 0.039 & 0.007 & 11.218 & 17.582 & 114.694 & 0.486 & 8.498 & 0.306 & 0.692 & 5.937 & 0.683 \\
            & $\pi^3$ & 0.200 & 0.082 & 0.010 & 6.034 & 11.387 & 34.416 & 0.227 & 6.173 & 0.626 & 1.068 & 1.881 & 0.690 \\
            & +Ours & \textbf{1.000} & \textbf{0.881} & \textbf{0.753} & \textbf{0.249} & \textbf{0.428} & \textbf{0.909} & \textbf{0.084} & \textbf{3.970} & \textbf{0.933} & \textbf{0.326} & \textbf{0.327} & \textbf{0.795} \\
            \midrule
            \multirow{3}{*}{\rotatebox{90}{\scriptsize S2D3DS}}
            & Wid3R & 0.260 & 0.191 & 0.035 & 2.912 & 3.374 & 31.397 & 0.364 & 1.416 & 0.309 & 0.455 & 0.446 & 0.623 \\
            & $\pi^3$ & 0.138 & 0.103 & 0.015 & 3.814 & 3.865 & 55.577 & 0.424 & 1.449 & 0.408 & 0.519 & 0.929 & 0.601 \\
            & +Ours & \textbf{0.833} & \textbf{0.767} & \textbf{0.525} & \textbf{1.022} & \textbf{1.202} & \textbf{11.347} & \textbf{0.075} & \textbf{0.469} & \textbf{0.954} & \textbf{0.170} & \textbf{0.160} & \textbf{0.787} \\
            \midrule
            \multirow{3}{*}{\rotatebox{90}{\scriptsize 360Loc}}
            & Wid3R & 0.500 & 0.243 & 0.097 & 4.452 & 5.457 & 9.092 & 0.414 & 7.390 & 0.285 & 1.110 & 0.750 & 0.627 \\
            & $\pi^3$ & 0.283 & 0.105 & 0.020 & 6.122 & 6.813 & 27.750 & 0.578 & 7.108 & 0.301 & 1.586 & 2.471 & 0.575 \\
            & +Ours & \textbf{0.995} & \textbf{0.960} & \textbf{0.806} & \textbf{0.220} & \textbf{0.266} & \textbf{1.098} & \textbf{0.286} & \textbf{4.713} & \textbf{0.815} & \textbf{0.469} & \textbf{0.372} & \textbf{0.720} \\
            \bottomrule[1pt]
        \end{tabular}
    }
    \caption{
        Quantitative evaluation on mixed-camera sequences.
    }
    \label{tab:quantitative_mixed}
\end{table*}

%% file: tab/quantitative_experts.tex
\begin{table*}[t]
    \centering
    \resizebox{\textwidth}{!}{
        \newcolumntype{M}{>{\centering\arraybackslash}p{7mm}}
        \begin{tabular}{l | *{7}{MM|} MM }
            \toprule[1pt]
            \multirow{3}{*}{Method} &
            \multicolumn{4}{c|}{Perspective} &
            \multicolumn{4}{c|}{Fisheye} &
            \multicolumn{4}{c|}{Panoramic} &
            \multicolumn{4}{c}{Mixed}
            \\
            & \multicolumn{2}{c|}{ADT} & \multicolumn{2}{c|}{ETH3D}
            & \multicolumn{2}{c|}{ADT} & \multicolumn{2}{c|}{ScanNet++}
            & \multicolumn{2}{c|}{S2D3DS} & \multicolumn{2}{c|}{360Loc}
            & \multicolumn{2}{c|}{S2D3DS} & \multicolumn{2}{c}{360Loc}
            \\
            & CD$\downarrow$ & NC$\uparrow$ & CD$\downarrow$ & NC$\uparrow$
            & CD$\downarrow$ & NC$\uparrow$ & CD$\downarrow$ & NC$\uparrow$
            & CD$\downarrow$ & NC$\uparrow$ & CD$\downarrow$ & NC$\uparrow$
            & CD$\downarrow$ & NC$\uparrow$ & CD$\downarrow$ & NC$\uparrow$
            \\ \midrule
            Original & \textbf{0.036} & \textbf{0.790} & \textbf{0.052} & \textbf{0.907} & 0.089 & 0.741 & 0.071 & 0.788 & 1.598 & 0.526 & 3.598 & 0.568 & 0.724 & 0.601 & 2.028 & 0.575 \\
            Finetune & 0.042 & 0.754 & 0.076 & 0.855 & 0.040 & 0.751 & 0.020 & 0.820 & 0.448 & 0.535 & 1.662 & 0.496 & 0.347 & 0.650 & 1.251 & 0.569 \\
            Retrain & 0.748 & 0.553 & 1.458 & 0.552 & 0.240 & 0.635 & 0.578 & 0.550 & 1.520 & 0.492 & 4.953 & 0.521 & 1.620 & 0.480 & 5.100 & 0.500 \\
            \midrule
            LoRA (3PFP) & 0.037 & 0.783 & 0.063 & 0.874 & 0.036 & 0.755 & \textbf{0.017} & 0.835 & \textbf{0.050} & 0.863 & 0.543 & 0.734 & 0.215 & 0.771 & 0.446 & 0.706 \\
            LoRA (1PFP) & 0.040 & 0.757 & 0.085 & 0.847 & 0.037 & 0.755 & 0.018 & 0.827 & 0.064 & 0.843 & 0.372 & 0.729 & 0.204 & 0.762 & 0.425 & 0.699 \\
            LoRA (1FP) & \textbf{0.036} & \textbf{0.790} & \textbf{0.052} & \textbf{0.907} & \textbf{0.035} & 0.752 & 0.018 & 0.834 & 0.063 & 0.837 & \textbf{0.344} & 0.739 & 0.178 & 0.770 & \textbf{0.415} & 0.707 \\
            LoRA (2FP) & \textbf{0.036} & \textbf{0.790} & \textbf{0.052} & \textbf{0.907} & \textbf{0.035} & \textbf{0.756} & \textbf{0.017} & \textbf{0.842} & 0.055 & \textbf{0.864} & 0.373 & \textbf{0.756} & \textbf{0.172} & 0.771 & 0.423 & \textbf{0.712} \\
            \midrule
            LoRA (2FP) Compressed & \textbf{0.036} & \textbf{0.790} & \textbf{0.052} & \textbf{0.907} & 0.036 & 0.750 & 0.019 & 0.818 & 0.064 & 0.845 & 0.393 & 0.740 & 0.179 & \textbf{0.772} & 0.435 & 0.701 \\
            \bottomrule[1pt]
        \end{tabular}
    }
    \caption{
        Quantitative point cloud evaluation with $\pi^3$ backbone.
        \textbf{3PFP}: 3 experts for perspective, fisheye, and panoramic images, respectively.
        \textbf{1PFP}: 1 expert for all images.
        \textbf{1FP}: 1 expert for only fisheye and panoramic images.
        \textbf{2FP}: 2 experts for fisheye and panoramic images, respectively.
    }
    \label{tab:quantitative_experts}
\end{table*}

%% file: fig/qualitative.tex
\begin{figure*}[t]
    \centering
    \includegraphics[width=\linewidth]{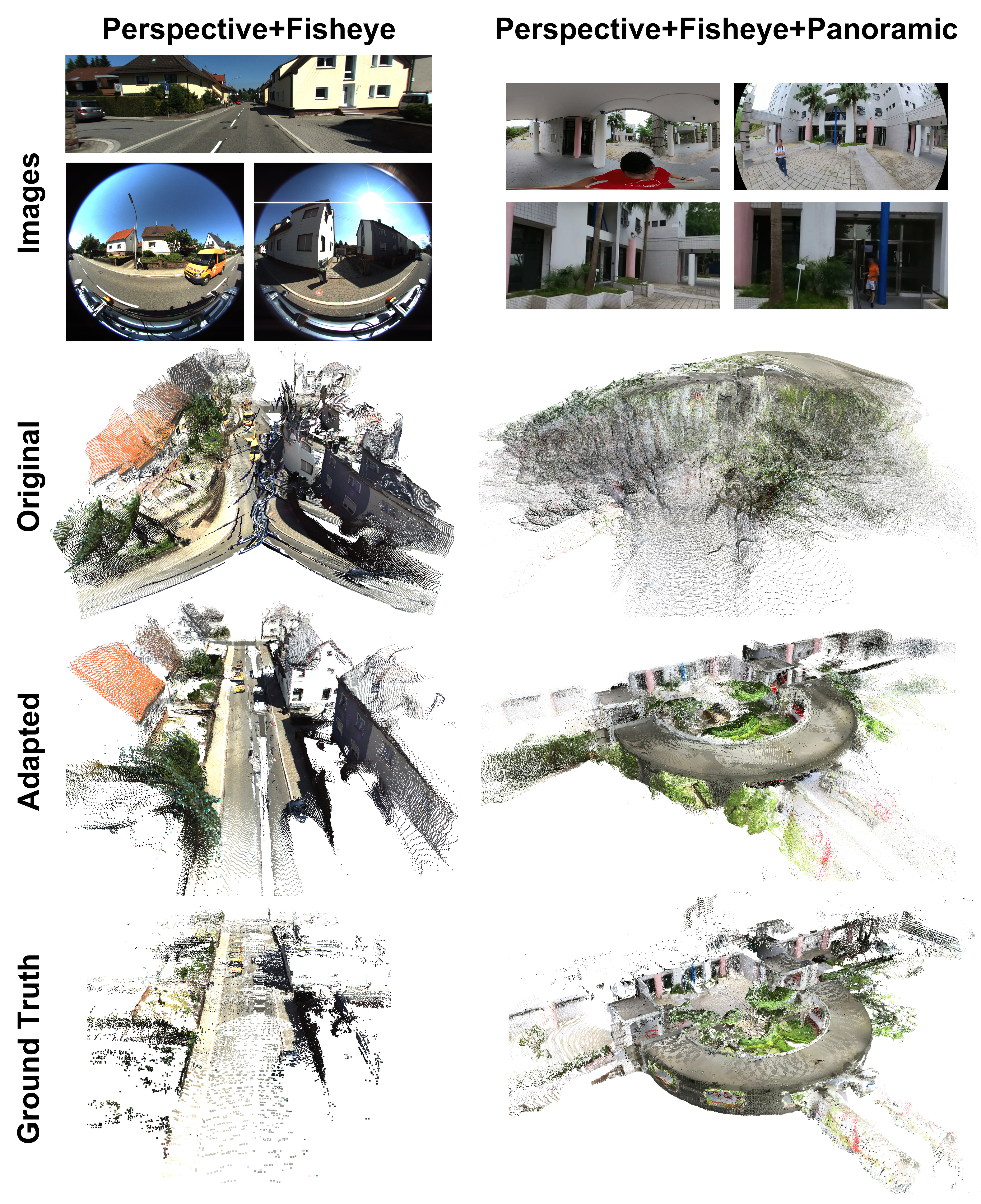}
    \caption{
        Qualitative reconstructions on mixed-camera sequences.
    }
    \label{fig:qualitative}
\end{figure*}

%% file: tab/datasets.tex
\begin{table}[ht]
    \centering
    \resizebox{0.9\textwidth}{!}{
        \begin{tabular}{cc|ccccc}
        \toprule[1pt]
            Split & Dataset & Camera & Domain & Environment \\
            \midrule
            \multirow{7}{*}{Train}
            & ScanNet++ \cite{yeshwanth2023scannet++} & Fisheye & Real & Indoor \\
            & ASE \cite{engel2023project} & Fisheye & Synthetic & Indoor \\
            & KITTI360 \cite{liao2022kitti} & Fisheye\&Perspective & Real & Outdoor \\
            & Matterport3D \cite{chang2017matterport3d} & Panoramic & Real & Indoor \\
            & Gibson \cite{xia2018gibson} & Panoramic & Real & Indoor \\
            & TartanAirV2 \cite{wang2020tartanair} & Panoramic & Synthetic & Indoor\&Outdoor \\
            & PanoCity \cite{guo2026panovggt} & Panoramic & Synthetic & Outdoor \\
            \midrule
            \multirow{5}{*}{Test}
            & ScanNet++ \cite{yeshwanth2023scannet++} & Fisheye & Real & Indoor \\
            & ADT \cite{pan2023aria} & Fisheye & Real\&Synthetic & Indoor \\
            & KITTI360 \cite{liao2022kitti} & Fisheye\&Perspective & Real & Outdoor \\
            & 360Loc \cite{huang2024360loc} & Panoramic & Real & Indoor\&Outdoor \\
            & Stanford2D3DS \cite{armeni2017joint} & Panoramic & Real & Indoor \\
        \bottomrule[1pt]
        \end{tabular}
    }
    \vspace{2mm}
    \caption{
        Non-perspective datasets used in training and evaluation.
    }
    \label{tab:datasets}
\end{table}

%% file: fig/vis_adt.tex
\begin{figure*}[p]
    \centering
    \includegraphics[width=\linewidth]{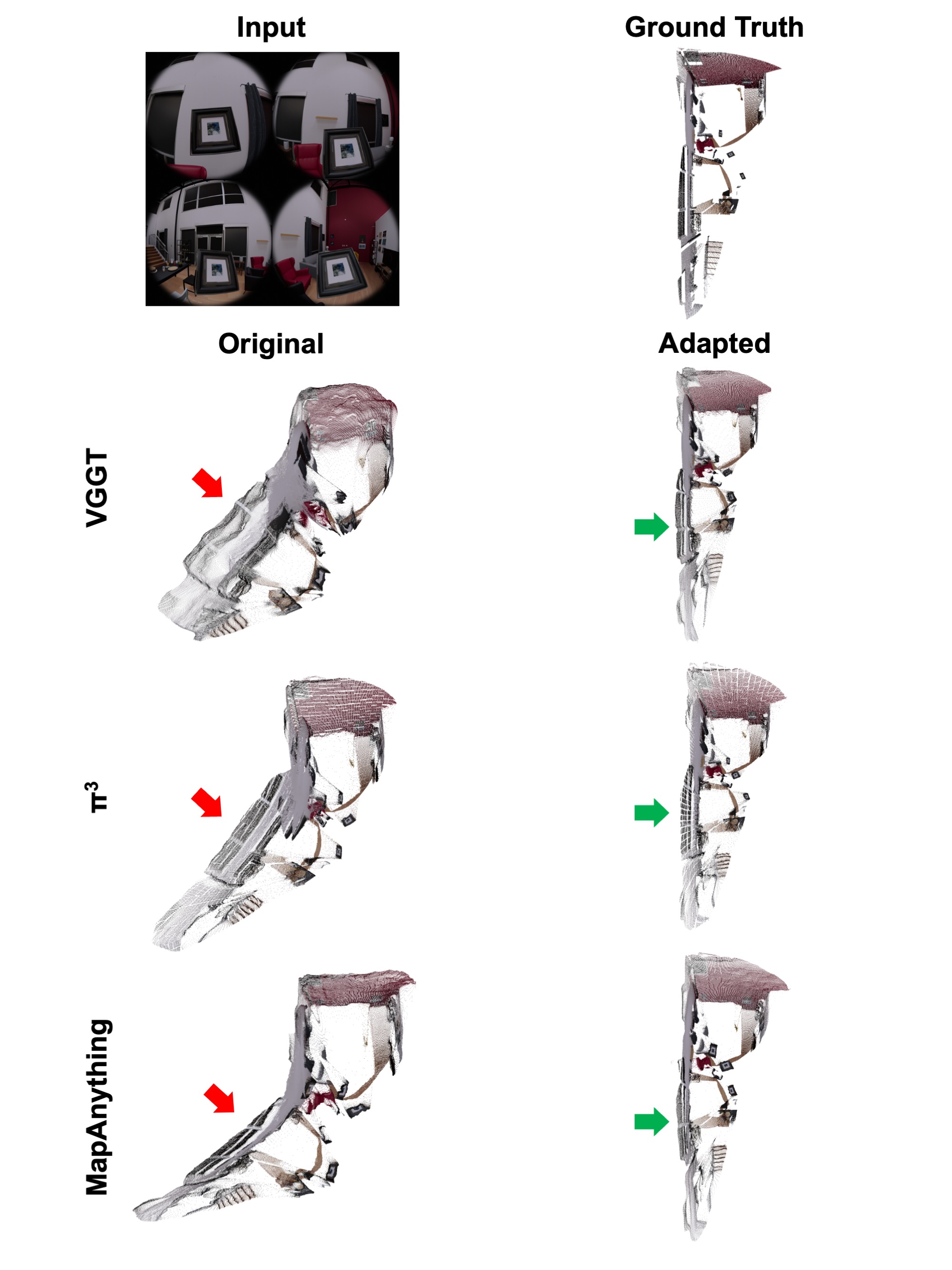}
    \caption{
        Qualitative reconstructions on ADT.
    }
    \label{fig:vis_adt}
\end{figure*}

%% file: fig/vis_scannetpp.tex
\begin{figure*}[p]
    \centering
    \includegraphics[width=\linewidth]{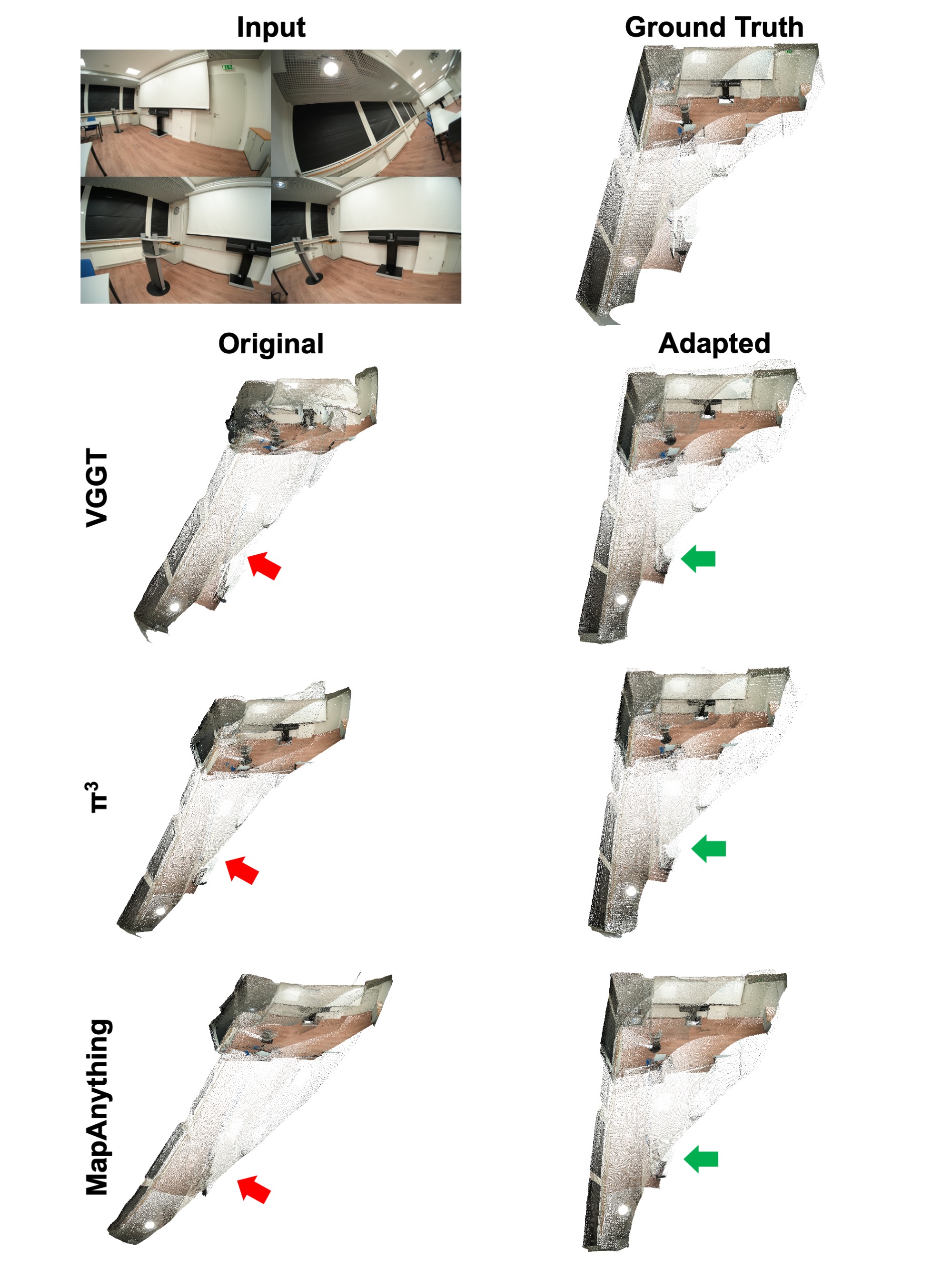}
    \caption{
        Qualitative reconstructions on ScanNet++.
    }
    \label{fig:vis_scannetpp}
\end{figure*}

%% file: fig/vis_stanford2d3ds.tex
\begin{figure*}[p]
    \centering
    \includegraphics[width=\linewidth]{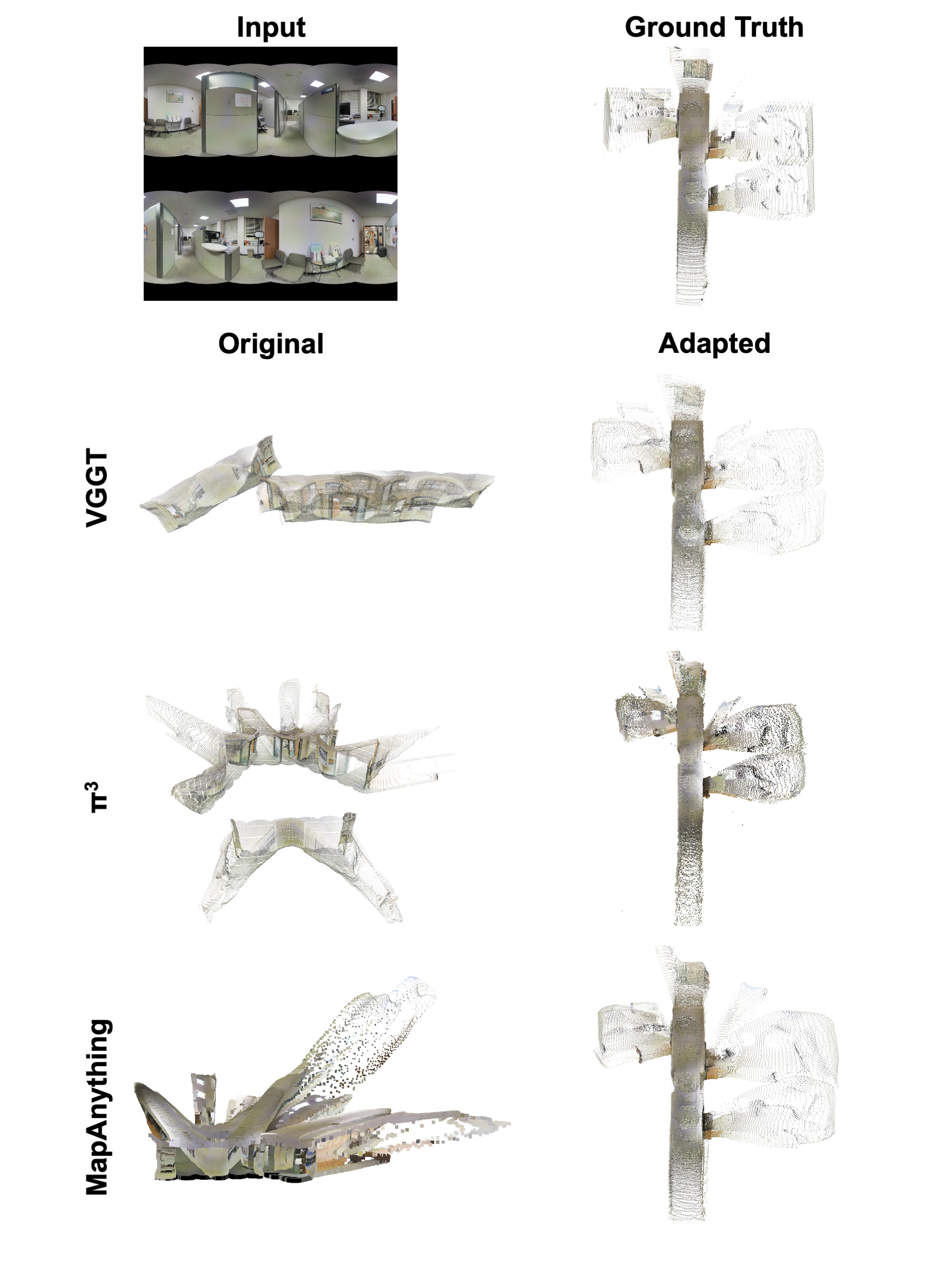}
    \caption{
        Qualitative reconstructions on Stanford2D3DS.
    }
    \label{fig:vis_stanford2d3ds}
\end{figure*}

%% file: fig/vis_360loc.tex
\begin{figure*}[p]
    \centering
    \includegraphics[width=\linewidth]{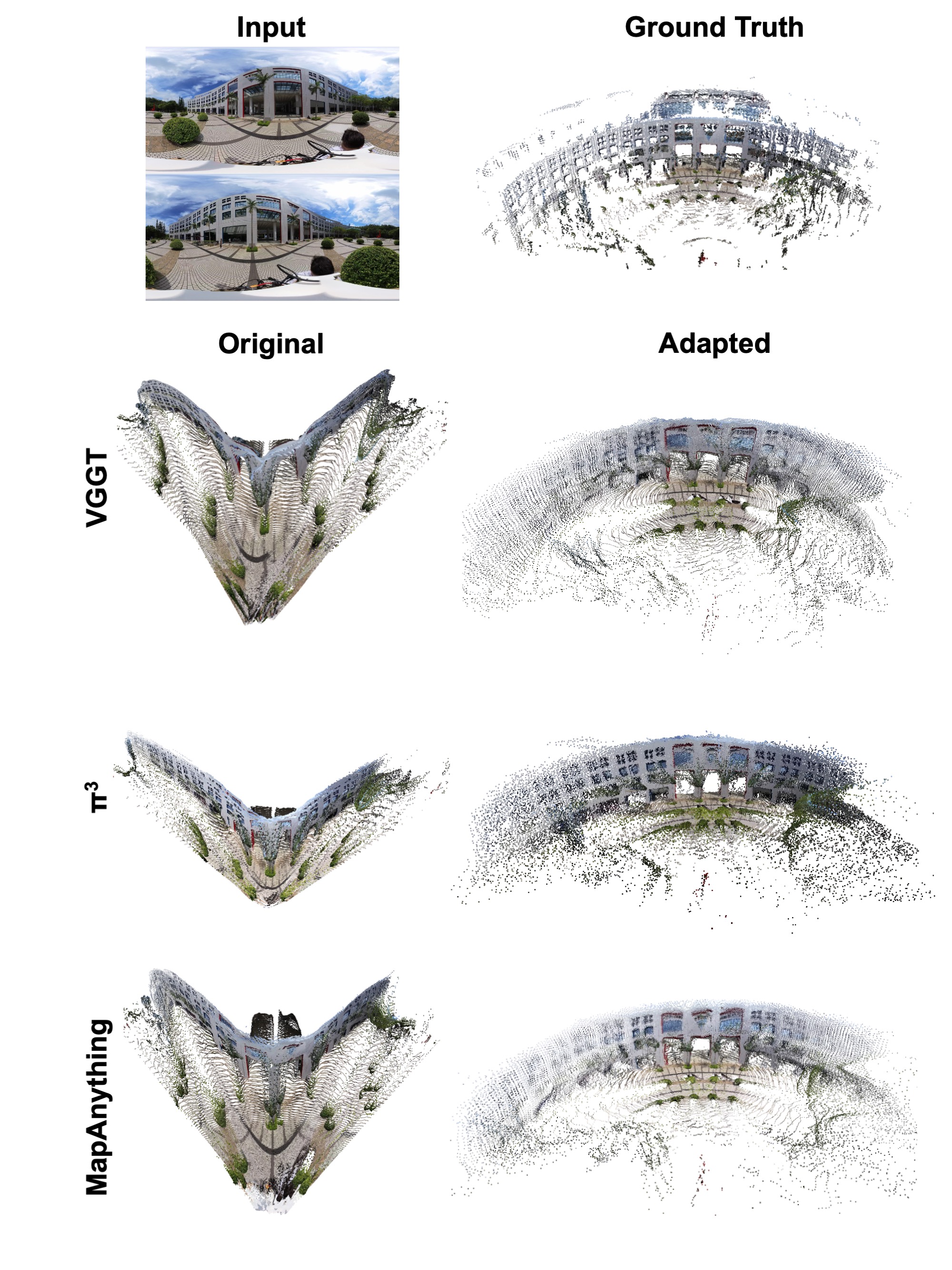}
    \caption{
        Qualitative reconstructions on 360Loc.
    }
    \label{fig:vis_360loc}
\end{figure*}